\documentclass[conference]{IEEEtran}
\IEEEoverridecommandlockouts
\usepackage{cite}
\usepackage{amsmath,amssymb,amsfonts}
\usepackage{makecell}  
\usepackage{graphicx}
\usepackage{textcomp}
\usepackage{booktabs}
\usepackage{array}
\usepackage{booktabs}
\usepackage{multirow}
\usepackage{float}
\usepackage{subcaption}
\usepackage{tcolorbox}
\usepackage{booktabs}
\usepackage{multirow}
\usepackage{caption}
\usepackage{graphicx}
\usepackage{graphicx}
\usepackage{svg}
\usepackage{graphicx}
\usepackage{svg}
\svgsetup{inkscapelatex=false} 
\usepackage{graphicx}
\usepackage{epstopdf}

\usepackage[a4paper, margin=1in]{geometry}
\usepackage{booktabs}
\usepackage{multirow}
\usepackage{caption}
\usepackage{subcaption}     
\usepackage{booktabs}
\usepackage{subcaption} 
\usepackage{multirow}
\usepackage{algorithm}
\usepackage{algpseudocode}
\usepackage{enumitem}

\makeatletter
\renewcommand{\theALG@line}{\arabic{ALG@line}}  

\makeatletter
\algrenewcommand\alglinenumber[1]{\footnotesize\makebox[1em][r]{#1}}
\makeatother
\newcolumntype{B}{>{\bfseries}c} 
\usepackage{xcolor}
\def\BibTeX{{\rm B\kern-.05em{\sc i\kern-.025em b}\kern-.08em
    T\kern-.1667em\lower.7ex\hbox{E}\kern-.125emX}}
\begin{document}

\title{AgentSME for Simulating Diverse Communication Modes in Smart Education\\
}

\author{
\IEEEauthorblockN{1\textsuperscript{st} Wen-Xi Yang}
\IEEEauthorblockA{
\textit{Guangdong Institute of Smart Education}\\
\textit{Jinan University}\\
Guangzhou, China
}
\and
\IEEEauthorblockN{2\textsuperscript{nd} Tian-Fang Zhao\textsuperscript{*}}
\IEEEauthorblockA{
\textit{School of Journalism \& Communication}\\
\textit{Jinan University}\\
Guangzhou, China\\
tianfang09@foxmail.com
}
}

\maketitle

\begin{abstract}
Generative agent models specifically tailored for smart education are critical, yet remain relatively underdeveloped. A key challenge stems from the inherent complexity of educational contexts: learners are human beings with various cognitive behaviors, and pedagogy is fundamentally centered on personalized human-to-human communication. To address this issue, this paper proposes AgentSME, a unified generative agent framework powered by LLM. Three directional communication modes are considered in the models, namely Solo, Mono, and Echo, reflecting different types of agency autonomy and communicative reciprocity. Accuracy is adopted as the primary evaluation metric, complemented by three diversity indices designed to assess the diversity of reasoning contents. Six widely used LLMs are tested to validate the robustness of communication modes across different model tiers, which are equally divided into base-capacity and high-capacity configurations. The results show that generative agents that employ the Echo communication mode achieve the highest accuracy scores, while DeepSeek exhibits the greatest diversity. This study provides valuable information to improve agent learning capabilities and inspire smart education models.
\end{abstract}


\section*{NOMENCLATURE}

\begin{description}[
    labelwidth=2cm,  
    labelsep=0.5em,  
    leftmargin=!      
]
    \item[$R$] The repeat count of independent experiments. Set $R = 10$ for each model.
    \item[$M$] The number of agents in each model. Set $M = 6$ in this paper.
    \item[$RC$] The content of the reasoning returned by LLMs. 
	\item[$Y$] The reasoning answer for each question in the datasets. 
	\item[$ASK$] The prompting words for LLM. 
	\item[$Q = \{q_k\}$] The question set for each independent experiment, with $k$ denotes the serial number of each question in their dataset. 
	\item[$A= \{a_i\}$] The set of agents for each independent experiment, with $i$ denotes the serial number of each agent. 
	\item[$Rlog$] The response log for each LLM response returned. 
    \item[$AgentSolo$] The generative agent model with solo-mode of communication. 
	\item[$AgentMono$] The generative agent model with mono-mode of communication. 
	\item[$AgentEcho$] The generative agent model with echo-mode of communication. 
	
\end{description}

\section{Introduction}
Smart education systems have emerged as a pivotal force in transforming traditional teaching paradigms. Conventional intelligent tutoring systems often rely on rule-based approaches \cite{nkambou2010advances} or static models \cite{graesser1999autotutor}, while flexibility to dynamically adapt to the needs of individual learners and provide interactive real-time experiences remains a concern. 

Recent years have witnessed growing interest in the application of generative models to education. Some early work primarily used generative adversarial networks and variational autoencoders for content generation \cite{kurdi2020systematic}, \cite{akkem2024comprehensive}, such as automated question creation and personalized learning materials \cite{cheng2024exploring}. Then, the emergence of large-scale pre-trained language models, including GPT and BERT variants, has significantly advanced the capabilities of generative agents to engage in human-like dialogue and adaptive tutoring \cite{liu2024personalized}. For example, previous studies have shown that generative tutors can provide personalized hints and guidance, helping students solve problems more effectively\cite{nye2023generative}. In parallel, some systems simulate peer learners to facilitate group learning and enhance student motivation\cite{zulfiqar2018using}. Building on these ideas, more recent works such as EduAgent\cite{xu2024eduagent} and Agent4Edu\cite{gao2025agent4edu} have begun to formalize the paradigm of applying AI agents in educational contexts, with the aim of supporting both individual tutoring and collaborative learning scenarios. The entities of generative agents powered by state-of-the-art generative models have demonstrated remarkable capabilities to simulate human-like behaviors \cite{park2023generative} and generate context-aware interactions \cite{zhang2021commentary}. These advances open new avenues for personalized education, where generative agents dynamically tailor instructional content and feedback based on the state and preferences of evolving knowledge \cite{kasneci2023chatgpt}, thus improving participation and learning outcomes. 

Despite the growing body of research on generative agents tailored for educational settings, relatively little attention has been paid to communication patterns between agents and how these patterns influence learning outcomes. In real world contexts, communication and interaction styles are highly diverse, ranging from lecture-based instruction to seminar-style discussions and self-directed learning \cite{liu2024classmeta}. These different pedagogical modes inherently shape the nature of the interaction between participants, which in turn affects both the engagement and the acquisition of knowledge \cite{chi2014icap}\cite{10472975}. However, most existing generative agent models overlook this important dimension, often assuming a homogeneous, one-size-fits-all dialogue structure. Moreover, current approaches focus predominantly on using large language models to simulate realistic behaviors of virtual students and tutors \cite{winkler2018unleashing}, with an emphasis on fidelity and immersion. While this contributes to more convincing simulations, it often neglects a fundamental perspective: agents themselves are learners. That is, generative agents, especially those acting as virtual tutors or peers, have inherent limitations in reasoning, adaptability, and selection of pedagogical strategies \cite{langley2009cognitive}. Enhancing the learning capabilities of the agents themselves, rather than only optimizing their states \cite{9036051}\cite{10897889}, has a significant potential to improve real student learning as well. 

Building on the observations above, this work proposes a novel framework of generative agents for smart education that explicitly incorporates the mode of communication as the core modeling factor. The proposed framework contributes to the field in three main ways.

First, a triadic taxonomy of communication modes (solo, mono, and echo) is designed to more faithfully reflect real-world educational interactions and learning behaviors. The solo mode simulates self-directed learning scenarios, where an agent independently performs iterative experiments and reflections to acquire knowledge. The mono mode captures unidirectional learning patterns, such as a student seeking guidance from a tutor or receiving direct instruction. In contrast, the echo mode simulates bidirectional peer-mode learning, in which two agents engage in mutual interaction and adaptive dialogue, leading to colearning and reciprocal improvement. This taxonomy enables a more granular analysis of how interaction dynamics influences the learning trajectories of generative agents.

Second, a comprehensive multidimensional evaluation framework is proposed to assess the learning capabilities of agents operating in different communication modes. Although most existing evaluations for generative agents rely on qualitative or non-deterministic text-based metric. This paper introduces a hybrid performance-centered metric approach. Specifically, accuracy is emphasized as the primary axis of evaluation, reflecting the correctness of an agent's responses. Three diversity-aware metrics are introduced, including Inverse Simpson to measure the dominance of high-frequency tokens, Honoré Statistic to quantify the contribution of low-frequency terms, and Information Entropy to evaluate the evenness of lexical distribution. These metrics provide a nuanced understanding of the expressive diversity of agent responses.

Third, extensive experiments are conducted across low-, medium-, and high-difficulty question datasets. Six prominent LLMs are tested for each communication mode. The results reveal that agents using the echo communication modes exhibit the highest gains in response accuracy, suggesting that mutual dialogue fosters deeper learning. DeepSeek demonstrates the highest lexical diversity across all difficulty levels, underscoring its potential for generating semantically rich responses.

\section{AgentSME}
AgentSME is a unified generative agent framework designed to simulate the influence of three classical communication modes on learning performance. The three modes, solo, mono, and echo, correspond respectively to self-directed learning, unidirectional instruction, and bidirectional peer interaction.
\subsection{Agent Role Setting}
In the AgentSME framework, each LLM is abstracted as a virtual student agent. To reflect heterogeneity among agents, the LLMs are divided into two distinct groups based on prior academic proficiency: a base-capacity group and a high-capacity group. The sizes of the two groups are balanced to ensure comparability and experimental validity. 
\begin{itemize}
\item High-capacity LLMs, which exhibit strong performance in both reasoning depth and knowledge coverage, including GPT-4o, DeepSeek-Reasoner, and Qwen-Plus. 
\item Base-capacity LLMs, consisting of representative yet lightweight models, including GPT-3.5-Turbo, DeepSeek-Chat, and Qwen-Turbo-Latest.
\end{itemize}

Different types of large language models exhibit varying capabilities and stylistic characteristics, often accompanied by inherent role configurations and hierarchical capability levels. For example, base-capacity LLMs are generally considered to have relatively lower reasoning depth and knowledge breadth compared to high-capacity LLMs. Moreover, model performance can vary across linguistic contexts, e.g., models trained or fine-tuned for Chinese, such as DeepSeek and Qwen, tend to outperform general-purpose models like ChatGPT when answering Chinese questions. This highlights the importance of language- and culture-specific adaptation in the evaluation and deployment of generative agents.

To avoid additional biases introduced by prompt engineering, all agents are tested using the same standardized format: single-choice questions. The presentation format is unified as follows: each question consists of a stem and four answer options. The agents outputs include chain-of-thought reasoning followed by a selected answer choice (A, B, C, or D). After answering each question, the context is reset to ensure that any performance variation can be fully attributed to the agent’s exploration and peer interactions within the current question.

\subsection{Communication Modes}
The mode of communication plays a crucial role in shaping the effectiveness of interactions among intelligent agents, particularly in educational contexts where collaborative learning and knowledge transfer are key objectives. For instance, Solo and Mono modes emphasize autonomous exploration, which can promote deeper individual reflection but may limit access to immediate peer feedback. In Mono mode, where one agent passively receives reasoning from others, knowledge delivery may be efficient, yet opportunities for interactive adaptation remain limited. In contrast, the Echo mode features bidirectional communication, where agents actively exchange their reasoning processes. This dynamic interaction allows them to clarify misunderstandings, resolve ambiguities, and co-construct knowledge collaboratively. Understanding and modeling these diverse communication modes is essential for designing generative agents with stronger reasoning capabilities and more adaptive learning behaviors.

1) Generative agent model with solo mode (AgentSolo). The solo-mode isolates each agent completely. The generative agent model receives only the question stem and four answer options and generates chain-of-thought reasoning followed by an immediate answer under fixed temperature and top-p sampling settings. Models using the solo strategy serve as baseline models to measure the inherent capabilities and output diversity of each model without external information. After answering each question, the context is immediately cleared to prevent information from leaking into subsequent questions.

\begin{algorithm}[t]
\caption{AgentSolo}
\label{alg:agent-s}

\textbf{Parameters:} repeat count $R = 10$; number of total agents $M = 6$;
reasoning content $RC$; answer $Y$; prompting words $ASK$.\\
\textbf{Input:} question set $Q=\{q_k\}$; learner set $A=\{a_i\}$.\\
\textbf{Output:} answer log $\mathcal{R}\!\log$.\\[0.4em]
\textbf{Procedure:}

{\small
\begin{algorithmic}[1]      
\For{$r \gets 1$ \textbf{to} $R$}
    \ForAll{$q_k \in Q$}
        \ForAll{$a_i \in A$}
            \State $ASK_i \gets BuildPrompt(a_i,q)$
            \State $(RC_i, Y_i) \gets GeneAnswer(a_i,ASK_i)$
            \State $\mathcal{R}\!\log \gets \mathcal{R}\!\log
                   \cup (r,q,a_i,Y_i,RC_i)$
        \EndFor
    \EndFor
\EndFor
\State \Return $\mathcal{R}\!\log$
\end{algorithmic}
}
\end{algorithm}

2) Generative Agent Model with Mono-mode (AgentMono). The mono style is designed to simulate scenarios of unidirectional consultation, inquiry, or instruction. The platform first randomly assigns a pair of roles, a mentor (the consulted agent), and a learner. The process consists of two steps. Firstly, the mentor independently generates and submits their chain-of-thought reasoning in response to a question. The learner then receives the complete reasoning text from the mentor, which is attached directly to the original prompt of the learner without additional instructions. The learner then reconsiders the problem and submits a final answer based on this information. The mentor’s answer remains unchanged throughout this process.
\begin{algorithm}[t]
\caption{AgentMono}
\label{alg:agent-m}

\textbf{Parameters:} repeat count $R = 10$; number of total agents $M = 6$; reasoning content of one agent $RC$; answer of one agent $Y$; prompting words $ASK$.\\
\textbf{Input:} question set $Q = \{q_k\}$; learner agent list $A = \{a_i\}$; mentor agent list $A = \{a_m\}$.\\
\textbf{Output:} answer log $\mathcal{R}\!\log$.\\[0.4em]
\textbf{Procedure:}
{\small
\begin{algorithmic}[1]
\For{$r \gets 1$ \textbf{to} $R$}
    \ForAll{$q_k \in Q$}
        \ForAll{$a_i \in A$}
            \ForAll{$a_m \in A$}
                \State $ASK_i \gets BuildPrompt(a_i, q)$
                \State $RC_m \gets GeneResponse(a_m, ASK_i)$
                \State $(RC_i, Y_i) \gets UpdateAnswer(a_i, RC_m)$
                \State $\mathcal{R}\!\log \gets \mathcal{R}\!\log \cup (r, q, a_i, a_m, Y_i, RC_i)$
            \EndFor
        \EndFor
    \EndFor
\EndFor
\State \Return $\mathcal{R}\!\log$
\end{algorithmic}
}

\end{algorithm}

3) Generative agent model with Echo mode (AgentEcho). The echo strategy is designed to model genuine peer-to-peer mutual assistance. The system randomly pairs two agents, each of whom first performs independent reasoning on a given question. Upon completion, they exchange their respective reasoning content. After the exchange, each agent incorporates the reasoning of the other in its own prompt and generates a new round of reasoning along with a revised response. Since both agents update their responses based on the input of the other, the echo strategy enables exploration of the upper bound of dialogue-based collaboration.

\begin{algorithm}[t]
\caption{AgentEcho}
\label{alg:agent-e}

\textbf{Parameters:} repeat count $R = 10$; number of total agents $M = 6$; reasoning content of one agent $RC$; answer of one agent $Y$; prompting words $ASK$.\\
\textbf{Input:} question set $Q = \{q_k\}$; learner agent list $A = \{a_i\}$.\\
\textbf{Output:} answer log $\mathcal{R}\!\log$.\\[0.4em]
\textbf{Procedure:}
{\small
\begin{algorithmic}[1]
\For{$r \gets 1$ \textbf{to} $R$}
    \ForAll{$q_k \in Q$}
        \ForAll{unordered pairs $(a_i, a_j) \in A$ where $a_i \ne a_j$}
            \State $ASK_i \gets BuildPrompt(a_i, q)$
            \State $ASK_j \gets BuildPrompt(a_j, q)$
            \State $RC_i \gets GeneResponse(a_i, ASK_j)$
            \State $RC_j \gets GeneResponse(a_j, ASK_i)$
            \State $(RC_i, Y_i) \gets UpdateAnswer(a_i, RC_j)$
            \State $(RC_j, Y_j) \gets UpdateAnswer(a_j, RC_i)$
            \State $\mathcal{R}\!\log \gets \mathcal{R}\!\log \cup (r, q, a_i, a_j, Y_i, Y_j, RC_i, RC_j)$
        \EndFor
    \EndFor
\EndFor
\State \Return $\mathcal{R}\!\log$
\end{algorithmic}

}

\end{algorithm}

\subsection{Evaluation Metrics}

To quantify the impact of the three communication styles on agent outputs, this study evaluates performance along two primary dimensions, accuracy, and linguistic diversity. 

1) Accuracy. It is strictly measured on the basis of the correctness of the given answers.

2) Linguistic diversity. It is further broken down into three complementary statistical metrics. 

\begin{itemize}
\item \textbf{Inverse Simpson Index.} The Inverse Simpson index is a statistical measure commonly used to quantify diversity, originally developed in ecological studies. In the context of this work, it evaluates linguistic diversity by considering both the richness (number of unique words) and the evenness (distribution balance) of word usage in the reasoning contents. The index is particularly sensitive to the dominance of high-frequency tokens and penalizes distributions where a small number of words appear excessively.

Formally, the Inverse Simpson is defined as:
\begin{equation}
Inverse\text{-}Simpson = \frac{N(N - 1)}{\sum_{v=1}^{V} \phi_v(\phi_v - 1)}
\end{equation}
where $N$ is the total number of tokens, $V$ is the number of unique word types, and $\phi_v$ is the frequency of the $v$-th word type. A higher Inverse-Simpson score indicates greater lexical diversity, suggesting that the word distribution is more balanced and that no small subset of tokens dominates the text.

\item \textbf{Honoré’s Statistic.} Honoré’s Statistic is a classical lexical diversity measure designed to capture the contribution of low-frequency words, particularly hapax legomena (words that occur only once in a text). Honoré’s Statistic is specifically tailored to highlight lexical richness at the long tail of the frequency spectrum, making it especially sensitive to rare or novel word usage. Honoré’s Statistic is computed as:
\begin{equation}
\text{Honoré-Statistic} = 100 \times \frac{\log N}{1 - \frac{V_1}{V}}
\end{equation}

Here, $N$ denotes the total number of tokens in the text, $V$ represents the number of unique word types, and $V_1$ refers to the number of word types that occur exactly once (i.e., \textit{hapax legomena}). A higher Honoré’s Statistic score indicates greater lexical diversity, especially when the text contains a larger proportion of rare or low-frequency words.

\item \textbf{Information Entropy.} It measures the uncertainty or unpredictability of a random variable, proposed by Claude Shannon in 1948. In this paper, it captures the lexical diversity of the reasoning content of LLMs by evaluating how evenly the vocabulary is distributed across generated outputs.
Formally, given a text with a vocabulary set $W = \{w_1, w_2, \dots, w_V\}$, the information entropy is computed as:

\begin{equation}
Information\text{-}Entropy = -\sum_{v=1}^{V} p(w_v) \log p(w_v)
\end{equation}

where $p(w_v)$ is the empirical probability of word $w_v$ occurring in the text. Higher Information-Entropy values indicate a more balanced and varied word distribution, suggesting richer language use, while lower Information-Entropy value reflects a concentration on fewer high-frequency terms.
\end{itemize}

\section{Experiments}

\subsection{Experiment Configuration}\label{AA}
\textbf{Datasets. }The experiments in this paper are carried out based on the society subset of the CMMLU dataset\cite{li2023cmmlu} . This subset contains a total of 264 single-choice questions covering fundamental concepts, classical theories, and research methods in sociology. To obtain the difficulty distribution of the questions and establish a baseline, this paper first conducts five independent reasoning rounds for each of the six models. All rounds use a unified prompt template and identical sampling hyperparameters (temperature = 0.7, top-p = 0.95). The questions in CMMLU are then categorized into three difficulty levels (HIGH, MEDIUM, LOW) based on the average error rate in the five rounds. The error rate is calculated as one minus the accuracy. 

\begin{itemize}
\item HIGH-difficulty questions: In five rounds of independent responses, the average error rate of LLM exceeded 80\%.
\item MEDIUM-difficulty questions: In five rounds of independent responses, the average error rate of the LLM ranged between 50\% and 80\%.
\item LOW-difficulty questions: In five rounds of independent responses, the average error rate of the LLM ranged between 20\% and 50\%.
\end{itemize}

Six LLMs (GPT-4o, DeepSeek-Reasoner, Qwen-Turbo-Latest, GPT-3.5-Turbo, DeepSeek-Chat, and Qwen-Plus) are used to simulate agents with different roles and learning abilities. Ten rounds of formal comparative testing are then conducted on each of the three difficulty level question sets based on the agents of Six LLMs, comparing the performance of the three communication modes in each round.

\begin{itemize}
\item Solo-mode: A single agent is randomly selected to answer questions independently, without access to peer information.
\item Mono-mode: A learner agent listens to mentor agents and re-answers the question based on the information received.
\item Echo-mode: Two agents powered by different LLMs are randomly selected to exchange their reasoning processes and then re-answer the question individually.
\end{itemize}

\subsection{Accuracy}
The results in Table \ref{tab:table1} demonstrate that, regardless of the level of the LLMs or the difficulty of the question, AgentEcho consistently outperforms AgentMono and AgentSolo, its accuracy advantage increasing as the difficulty of the task increases.

\begin{table*}[htbp] 
\centering
\caption{Performance Metrics of Generative Agents Across Datasets}
\label{tab:table1}

\begin{subtable}[t]{\linewidth}
\caption{For High-Capacity LLMs}
\centering
\begin{tabular}{p{1.2cm}p{1cm}p{1cm}p{1cm}p{1cm}p{1cm}p{1cm}p{1cm}p{1cm}p{1cm}p{1cm}}
\toprule
\textbf{Dataset} & \textbf{Metric} & \multicolumn{3}{c}{\textbf{Qwen-Plus}} & \multicolumn{3}{c}{\textbf{GPT-4o}} & \multicolumn{3}{c}{\textbf{Deepseek-Reasoner}} \\
\cmidrule(r){3-5} \cmidrule(r){6-8} \cmidrule(r){9-11}
& & \textbf{Solo} & \textbf{Mono} & \textbf{Echo} & \textbf{Solo }& \textbf{Mono} & \textbf{Echo} & \textbf{Solo }& \textbf{Mono} & \textbf{Echo} \\
\midrule

\multirow{4}{*}{\textbf{HIGH}}
& Mean     & 0.120 & 0.055 & \textbf{0.587} & 0.035 & 0.059 & \textbf{0.367} & 0.120 & 0.162 & \textbf{0.502} \\
& Best     & 0.150 & 0.100 & 0.800 & 0.150 & 0.150 & 0.800 & 0.150 & 0.300 & 0.800 \\
& Std      & 0.026 & 0.034 & 0.095 & 0.047 & 0.040 & 0.149 & 0.026 & 0.052 & 0.162 \\
& P\_value & 0.002** & 0.002** & \multicolumn{1}{c}{/} & 0.002** & 0.002** & \multicolumn{1}{c}{/} & 0.002** & 0.002** & \multicolumn{1}{c}{/} \\
\midrule

\multirow{4}{*}{\textbf{MEDIUM}}
& Mean     & 0.395 & 0.443 & \textbf{0.849} & 0.185 & 0.262 & \textbf{0.627} & 0.684 & 0.618 & \textbf{0.799}  \\
& Best     & 0.430 & 0.520 & 1.000 & 0.240 & 0.430 & 0.950 & 0.760 & 0.760 & 1.000 \\
& Std      & 0.024 & 0.042 & 0.068 & 0.044 & 0.058 & 0.144 & 0.058 & 0.066 & 0.109 \\
& P\_value & 0.002** & 0.002** & \multicolumn{1}{c}{/} & 0.002** & 0.002** & \multicolumn{1}{c}{/} & 0.002** & 0.002*** & \multicolumn{1}{c}{/} \\
\midrule

\multirow{4}{*}{\textbf{LOW}}
& Mean     & 0.857 & 0.863 & \textbf{0.958} & 0.630 & 0.679 & \textbf{0.877} & 0.786 & 0.860 & \textbf{0.935} \\
& Best     & 0.890 & 0.930 & 1.000 & 0.700 & 0.800 & 1.000 & 0.890 & 0.950 & 1.000 \\
& Std      & 0.015 & 0.026 & 0.029 & 0.053 & 0.053 & 0.082 & 0.062 & 0.040 & 0.052 \\
& P\_value & 0.002** & 0.002** & \multicolumn{1}{c}{/} & 0.002** & 0.002** & \multicolumn{1}{c}{/} & 0.002** & 0.002** & \multicolumn{1}{c}{/} \\
\bottomrule
\end{tabular}
\end{subtable}

\vspace{1em}

\begin{subtable}[t]{\textwidth}
\caption{For Base-capacity LLMs}
\centering

\begin{tabular}{p{1cm}p{1cm}p{1cm}p{1cm}p{1cm}p{1cm}p{1cm}p{1cm}p{1cm}p{1cm}p{1cm}}
\toprule
\textbf{Dataset} & \textbf{Metric} & \multicolumn{3}{c}{\textbf{Qwen-Turbo-Latest}} & \multicolumn{3}{c}{\textbf{GPT-3.5-Turbo}} & \multicolumn{3}{c}{\textbf{Deepseek-Chat}} \\
\cmidrule(r){3-5} \cmidrule(r){6-8} \cmidrule(r){9-11}
& & \textbf{Solo} & \textbf{Mono} & \textbf{Echo} & \textbf{Solo} & \textbf{Mono }& \textbf{Echo} & \textbf{Solo} & \textbf{Mono} & \textbf{Echo} \\
\midrule

\multirow{4}{*}{\textbf{HIGH}}
& Mean     & 0.050 & 0.051 & \textbf{0.521} & 0.085 & 0.108 & \textbf{0.284} & 0.040 & 0.050 & \textbf{0.394} \\
& Best     & 0.050 & 0.100 & 0.800 & 0.100 & 0.200 & 0.650 & 0.050 & 0.050 & 0.700 \\
& Std      & 0.000 & 0.005 & 0.187 & 0.024 & 0.038 & 0.133 & 0.021 & 0.000 & 0.167 \\
& P\_value & 0.002** & 0.002** & \makecell[c]{/} & 0.002* & 0.002** & \makecell[c]{/} & 0.002** & 0.002** & \makecell[c]{/} \\
\midrule

\multirow{4}{*}{\textbf{MEDIUM}}
& Mean     & 0.107 & 0.092 & \textbf{0.747} & 0.205 & 0.156 & \textbf{0.475} & 0.580 & 0.586 & \textbf{0.806} \\
& Best     & 0.140 & 0.190 & 0.950 & 0.240 & 0.240 & 0.810 & 0.620 & 0.620 & 1.000 \\
& Std      & 0.028 & 0.042 & 0.164 & 0.024 & 0.051 & 0.147 & 0.021 & 0.024 & 0.108 \\
& P\_value & 0.002** & 0.002** & \makecell[c]{/} & 0.002** & 0.002** & \makecell[c]{/} & 0.002** & 0.002** & \makecell[c]{/} \\
\midrule

\multirow{4}{*}{\textbf{LOW}}
& Mean     & 0.504 & 0.509 & \textbf{0.896} & 0.218 & 0.321 & \textbf{0.570} & 0.766 & 0.788 & \textbf{0.937} \\
& Best     & 0.520 & 0.550 & 1.000 & 0.300 & 0.390 & 0.890 & 0.800 & 0.860 & 1.000 \\
& Std      & 0.008 & 0.019 & 0.108 & 0.034 & 0.033 & 0.135 & 0.020 & 0.029 & 0.054 \\
& P\_value & 0.002** & 0.002** & \makecell[c]{/} & 0.002** & 0.002** & \makecell[c]{/} & 0.002** & 0.002** & \makecell[c]{/}\\
\bottomrule
\end{tabular}
\end{subtable}
\end{table*}

For HIGH-difficulty questions (20 items): Across agents of the six LLMs, the average accuracy improves from $0.035$–$0.120$ for \textbf{AgentSolo} to $0.284$–$0.587$ of \textbf{AgentEcho}, yielding gains of $0.249$–$0.467$ ($p < 0.01$). The most notable improvements are observed in \textsc{Qwen-Turbo-Latest}, which increased from $0.050$ to $0.521$. In contrast, \textbf{AgentMono} provided only marginal improvements ($\leq 0.001$), indicating that the \textbf{AgentMono} is insufficient to correct reasoning errors in high-difficulty tasks.

For MEDIUM-difficulty questions (21 items): The average accuracy of \textbf{AgentSolo} in MEDIUM-difficulty questions ($0.107$-$0.684$) performs better than that in HIGH-difficulty questions ($0.035$–$0.120$). Without exception, the average accuracy of \textbf{AgentEcho} in MEDIUM-difficulty questions ($0.475$-$0.849$) performs much better than in HIGH-difficulty questions ($0.284$–$0.587$).

For LOW-difficulty questions (44 items): \textbf{AgentSolo} achieves an accuracy of $0.218$-$0.857$. In this setting, \textbf{AgentEcho} achieved a high accuracy ($0.570$-$0.958$). \textbf{AgentMono}'s performance is very close to the \textbf{AgentSolo} model.

The overall effectiveness ranking is \textbf{AgentEcho} $>$ \textbf{AgentMono} $\approx$ \textbf{AgentSolo}. The echo communication mode has been shown to be an effective strategy for improving performance in all problems. It is worth noting that \textbf{AgentEcho} introduces greater round-to-round variability on hard questions, with increases in standard deviations of up to $0.187$, whereas \textbf{AgentMono} maintains a variance level similar to \textbf{AgentSolo}.

\subsection{Linguistic Diversity}
\begin{figure*}[htbp]
\centering
\begin{subfigure}[b]{0.60\textwidth}
    \includegraphics[width=\linewidth]{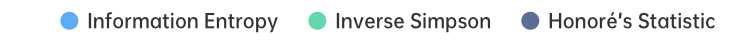}
    \label{fig:top1}
\end{subfigure}
\vspace{0.1em}  

\begin{subfigure}[b]{0.32\textwidth}
    \includegraphics[width=\linewidth]{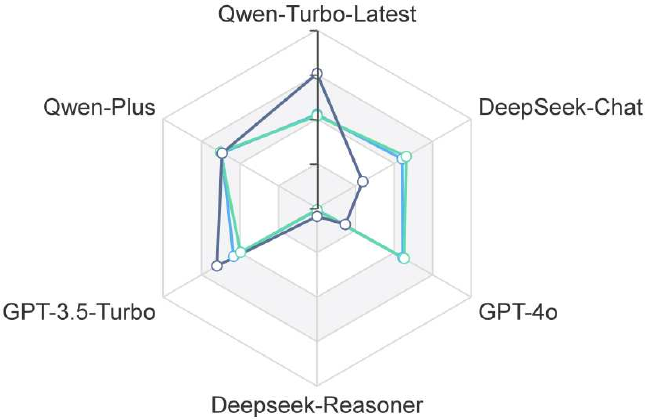}
    \caption{For AgentSolo}
    \label{fig:2a}
\end{subfigure}
\hfill
\begin{subfigure}[b]{0.32\textwidth}
    \includegraphics[width=\linewidth]{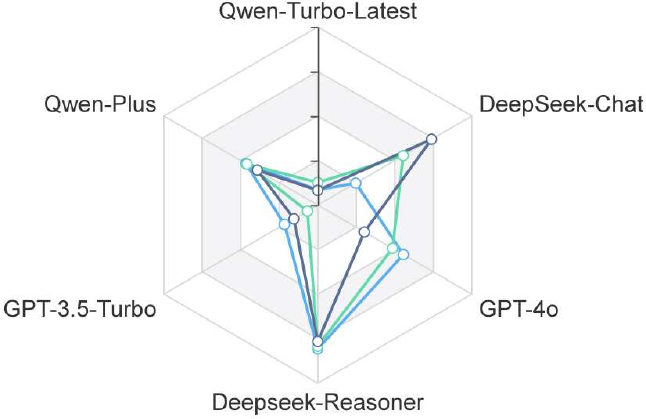}
    \caption{For AgentMono}
    \label{fig:2b}
\end{subfigure}
\hfill
\begin{subfigure}[b]{0.32\textwidth}
    \includegraphics[width=\linewidth]{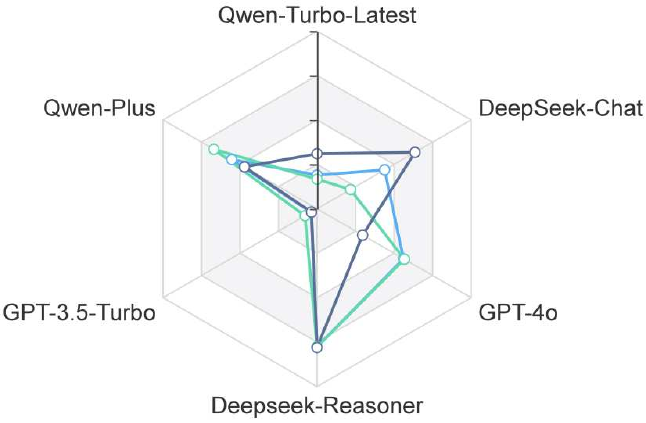}
    \caption{For AgentEcho}
    \label{fig:2c}
\end{subfigure}

\caption{Radar charts of six LLMs’ diversity across different modes and metrics.}
\label{fig:double-row-radar}
\end{figure*}

Fig.\ref{fig:double-row-radar} shows radar charts of the lexical diversity exhibited by six LLMs under three interaction paradigms: Solo (Fig.~1(a)), Mono  (Fig.~1(b)), and Echo  (Fig.~1(a)).  Each radius corresponds to one of three metrics (Inverse Simpson, Honoré’s statistic, and information entropy).  To allow direct comparison, all raw scores were first standardized within each mode via z–score normalization and then transformed using the standard normal cumulative distribution function into the range [0, 1] (where 1.0 represents the highest observed diversity and 0.0 the lowest).

As illustrated in Fig.1(a)–(c), the radar charts reveal clear shifts in lexical diversity as interaction depth increases across the Solo, Mono, and Echo communication modes. In Solo mode, where models operate independently, Qwen-Plus exhibits the most balanced and outward-reaching profile across all three metrics, indicating strong baseline diversity. In contrast, DeepSeek-Reasoner collapses near the center, reflecting limited diversity in the absence of interaction. GPT-3.5-Turbo and Qwen-Turbo-Latest also present highly contracted shapes, with especially low values on the entropy and Honoré axes, suggesting weak generative variation under isolated conditions.

In Mono mode, where each model receives only one round of feedback, diversity trajectories diverge. Qwen-Plus continues to maintain a well-balanced and expansive profile, demonstrating strong adaptability to limited interaction. Interestingly, DeepSeek-Reasoner undergoes a sharp improvement, with all three points extending outward—suggesting that even a single round of peer feedback partially activates its long-tail generative capacity. Conversely, Qwen-Turbo-Latest shows significant regression, especially in entropy and Honoré’s statistic, implying poor responsiveness to external signals. GPT-3.5-Turbo remains contracted, while GPT-4o and DeepSeek-Chat maintain moderate, stable profiles.

With full multi-turn interaction in Echo mode, DeepSeek-Reasoner achieves the most expansive and symmetrical radar shape, with all metrics reaching the outermost bounds—indicating full activation of its generative diversity potential. Qwen-Plus remains competitive, maintaining a high-diversity profile across metrics. DeepSeek-Chat also expands its reach, particularly on entropy and Honoré axes, showing its ability to incorporate collaborative input. Meanwhile, GPT-4o and Qwen-Turbo-Latest exhibit modest improvements, and GPT-3.5-Turbo stays near the center, confirming its limited responsiveness to extended feedback.

These findings underscore that lexical diversity does not correlate linearly with accuracy, and its variation depends heavily on interaction design and model architecture. While high-capacity models like GPT-4o deliver consistent performance, their lexical variety remains stable, suggesting a ceiling effect. In contrast, models like DeepSeek-Reasoner and Qwen-Plus demonstrate both high accuracy and enhanced diversity in Echo mode, reflecting their capacity to utilize peer interaction effectively. Overall, Echo interaction proves most beneficial for weaker or more adaptable models, providing not only accuracy gains but also richer, more varied language generation.

\section*{Conclusion}
This paper presents AgentSME, a generative agent framework for simulating multiple communication modes in smart education scenarios. By modeling Solo, Mono, and Echo modes through large language model (LLM)-based agents, the framework provides a scalable approach for analyzing how different interaction patterns influence learning performance and linguistic diversity.

Extensive experiments on the society subset of the CMMLU dataset—across three difficulty levels and six representative LLMs—demonstrate the superior effectiveness of the AgentEcho model. Not only does Echo mode significantly enhance answer accuracy, especially on high-difficulty questions, but it also fosters more diverse and expressive language generation. Meanwhile, AgentSolo proves to be the most cost-effective option for low-difficulty tasks, and AgentMono shows moderate gains but limited impact on diversity.

These findings reveal a non-linear relationship between accuracy and linguistic diversity,and highlight the unique value of multi-turn interaction. Echo-based collaboration is particularly beneficial for weaker models, supporting both reasoning correction and stylistic variation. This suggests promising directions for leveraging peer-style interaction to improve AI-assisted learning environments.
\section*{Acknowledgment}
This work is supported in part by the National Natural Science Foundation of China under Grant No. 62206112, and in part by the Humanities and Social Sciences Project of the Ministry of Education of China under Grant No. 22YJC860037, and in part by the Key Laboratory of Smart Education of Guangdong Higher Education Institutes under Grant No. 2022LSYS003. (Corresponding author: Tian-Fang Zhao).

\bibliographystyle{IEEEtran}
\bibliography{Mybib}

\begin{thebibliography}{10}

\bibitem{nkambou2010advances}
Roger Nkambou, Riichiro Mizoguchi, and Jacqueline Bourdeau.
\newblock {\em Advances in intelligent tutoring systems}, volume 308.
\newblock Springer, 2010.

\bibitem{graesser1999autotutor}
Arthur~C Graesser, Katja Wiemer-Hastings, Peter Wiemer-Hastings, Roger Kreuz, Tutoring~Research Group, et~al.
\newblock Autotutor: A simulation of a human tutor.
\newblock {\em Cognitive Systems Research}, 1(1):35--51, 1999.

\bibitem{kurdi2020systematic}
Ghader Kurdi, Jared Leo, Bijan Parsia, Uli Sattler, and Salam Al-Emari.
\newblock A systematic review of automatic question generation for educational purposes.
\newblock {\em International Journal of Artificial Intelligence in Education}, 30:121--204, 2020.

\bibitem{akkem2024comprehensive}
Yaganteeswarudu Akkem, Saroj~Kumar Biswas, and Aruna Varanasi.
\newblock A comprehensive review of synthetic data generation in smart farming by using variational autoencoder and generative adversarial network.
\newblock {\em Engineering Applications of Artificial Intelligence}, 131:107881, 2024.

\bibitem{cheng2024exploring}
Yuheng Cheng, Ceyao Zhang, Zhengwen Zhang, Xiangrui Meng, Sirui Hong, Wenhao Li, Zihao Wang, Zekai Wang, Feng Yin, Junhua Zhao, et~al.
\newblock Exploring large language model based intelligent agents: Definitions, methods, and prospects.
\newblock {\em arXiv preprint arXiv:2401.03428}, 2024.

\bibitem{liu2024personalized}
Dejian Liu, Ronghuai Huang, Ying Chen, Michael~Agyemang Adarkwah, Xiangling Zhang, Xin Li, Junjie Zhang, and Ting Da.
\newblock Personalized tutoring through conversational agents.
\newblock In {\em Using Educational Robots to Enhance Learning: An Analysis of 100 Academic Articles}, pages 59--85. Springer, 2024.

\bibitem{nye2023generative}
Benjamin~D Nye, Dillon Mee, and Mark~G Core.
\newblock Generative large language models for dialog-based tutoring: An early consideration of opportunities and concerns.
\newblock In {\em LLM@ AIED}, pages 78--88, 2023.

\bibitem{zulfiqar2018using}
Salman Zulfiqar, Rongting Zhou, Fahad Asmi, and Affan Yasin.
\newblock Using simulation system for collaborative learning to enhance learner’s performance.
\newblock {\em Cogent Education}, 5(1):1424678, 2018.

\bibitem{xu2024eduagent}
Songlin Xu, Xinyu Zhang, and Lianhui Qin.
\newblock Eduagent: Generative student agents in learning.
\newblock {\em arXiv preprint arXiv:2404.07963}, 2024.

\bibitem{gao2025agent4edu}
Weibo Gao, Qi~Liu, Linan Yue, Fangzhou Yao, Rui Lv, Zheng Zhang, Hao Wang, and Zhenya Huang.
\newblock Agent4edu: Generating learner response data by generative agents for intelligent education systems.
\newblock {\em arXiv preprint arXiv:2501.10332}, 2025.

\bibitem{park2023generative}
Joon~Sung Park, Joseph O'Brien, Carrie~Jun Cai, Meredith~Ringel Morris, Percy Liang, and Michael~S Bernstein.
\newblock Generative agents: Interactive simulacra of human behavior.
\newblock In {\em Proceedings of the 36th annual acm symposium on user interface software and technology}, pages 1--22, 2023.

\bibitem{zhang2021commentary}
Min Zhang and Juntao Li.
\newblock A commentary of gpt-3 in mit technology review 2021.
\newblock {\em Fundamental Research}, 1(6):831--833, 2021.

\bibitem{kasneci2023chatgpt}
Enkelejda Kasneci, Kathrin Se{\ss}ler, Stefan K{\"u}chemann, Maria Bannert, Daryna Dementieva, Frank Fischer, Urs Gasser, Georg Groh, Stephan G{\"u}nnemann, Eyke H{\"u}llermeier, et~al.
\newblock Chatgpt for good? on opportunities and challenges of large language models for education.
\newblock {\em Learning and individual differences}, 103:102274, 2023.

\bibitem{liu2024classmeta}
Ziyi Liu, Zhengzhe Zhu, Lijun Zhu, Enze Jiang, Xiyun Hu, Kylie~A Peppler, and Karthik Ramani.
\newblock Classmeta: Designing interactive virtual classmate to promote vr classroom participation.
\newblock In {\em Proceedings of the 2024 CHI Conference on Human Factors in Computing Systems}, pages 1--17, 2024.

\bibitem{chi2014icap}
Michelene~TH Chi and Ruth Wylie.
\newblock The icap framework: Linking cognitive engagement to active learning outcomes.
\newblock {\em Educational psychologist}, 49(4):219--243, 2014.

\bibitem{10472975}
Ming Gu, Tian-Fang Zhao, Liang Yang, Xiao-Kun Wu, and Wei-Neng Chen.
\newblock Modeling information cocoons in networked populations: Insights from backgrounds and preferences.
\newblock {\em IEEE Transactions on Computational Social Systems}, 11(3):4497--4510, 2024.

\bibitem{winkler2018unleashing}
Rainer Winkler and Matthias S{\"o}llner.
\newblock Unleashing the potential of chatbots in education: A state-of-the-art analysis.
\newblock In {\em Academy of management proceedings}, volume 2018, page 15903. Academy of Management Briarcliff Manor, NY 10510, 2018.

\bibitem{langley2009cognitive}
Pat Langley, John~E Laird, and Seth Rogers.
\newblock Cognitive architectures: Research issues and challenges.
\newblock {\em Cognitive Systems Research}, 10(2):141--160, 2009.

\bibitem{9036051}
Tian-Fang Zhao, Wei-Neng Chen, Sam Kwong, Tian-Long Gu, Hua-Qiang Yuan, Jie Zhang, and Jun Zhang.
\newblock Evolutionary divide-and-conquer algorithm for virus spreading control over networks.
\newblock {\em IEEE Transactions on Cybernetics}, 51(7):3752--3766, 2021.

\bibitem{10897889}
Ming Gu, Tian-Fang Zhao, Jinghui Zhong, and Wei-Neng Chen.
\newblock Progressive community merging cooperative coevolution algorithm for influence blocking maximization in social networks.
\newblock {\em IEEE Transactions on Network Science and Engineering}, 12(3):2093--2106, 2025.

\bibitem{li2023cmmlu}
Haonan Li, Yixuan Zhang, Fajri Koto, Yifei Yang, Hai Zhao, Yeyun Gong, Nan Duan, and Timothy Baldwin.
\newblock Cmmlu: Measuring massive multitask language understanding in chinese.
\newblock {\em arXiv preprint arXiv:2306.09212}, 2023.

\end{thebibliography}

\end{document}